\documentclass[letterpaper,twocolumn,fleqn]{article} 

\usepackage{ist}
\usepackage{comment}
\usepackage{amsmath,graphicx}
\usepackage{multirow}
\usepackage{caption}
\usepackage{booktabs}
\usepackage{cite} 
\usepackage{setspace}
\usepackage{adjustbox} 
\usepackage{subfigure}
\usepackage{array}
\usepackage{xspace}
\usepackage{graphicx}
\usepackage{amsmath,amssymb} 
\usepackage{lipsum}
\usepackage{xcolor}
\usepackage[switch]{lineno}
\usepackage{makecell}

\newcommand\blfootnote[1]{%
  \begingroup
  \renewcommand\thefootnote{}\footnote{#1}%
  \addtocounter{footnote}{-1}%
  \endgroup
}

\title{Can Image Enhancement be Beneficial to Find Smoke Images in Laparoscopic Surgery?}

\author{Congcong Wang $^{1,\star}$, Vivek Sharma $^{2,\star}$, Yu Fan$^{1}$, Faouzi Alaya Cheikh $^{1}$, Azeddine Beghdadi $^{3}$, Ole Jacob Elle $^{4,5}$, and Rainer Stiefelhagen $^{2}$\\
$^{1}$ Norwegian University of Science \& Technology, Norway. $^{2}$ Karlsruhe Institute of Technology, Germany.   $^{3}$ University Paris 13, France. $^{4}$ Oslo University Hospital, Norway. $^{5}$ University of Oslo, Norway.\\ \tt\small   \{congcong.wang,yu.fan,faouzi.cheikh\}@ntnu.no, \{vivek.sharma,rainer.stiefelhagen\}@kit.edu,  oelle@ous-hf.no,  azeddine.beghdadi@univ-paris13.fr}
\date{28th Feb., 2017} 

\date{} 

\begin{document} 

\maketitle 

\thispagestyle{empty} 


\begin{abstract}
Laparoscopic surgery has a limited field of view. Laser ablation in a laproscopic surgery causes smoke, which inevitably influences the surgeon's visibility. Therefore, it is of vital importance to remove the smoke, such that a clear visualization is possible. In order to employ a desmoking technique, one needs to know beforehand if the image contains smoke or not, to this date, there exists no accurate method that could classify the smoke/non-smoke images completely. In this work, we propose a new  enhancement method which enhances the informative details in the RGB images for discrimination of smoke/non-smoke images. Our proposed method utilizes weighted least squares optimization framework~(WLS). For feature extraction, we use statistical features based on bivariate histogram distribution of gradient magnitude~(GM) and Laplacian of Gaussian~(LoG). We then train a SVM classifier with binary smoke/non-smoke classification task. We demonstrate the effectiveness of our method on Cholec80 dataset. Experiments using our proposed enhancement method show promising results with improvements of 4\% in accuracy and 4\% in F1-Score over the baseline performance of RGB images. In addition, our approach improves over the saturation histogram based classification methodologies Saturation Analysis~(SAN) and Saturation Peak Analysis~(SPA) by 1/5\% and 1/6\% in accuracy/F1-Score metrics.

We can employ our enhancement method in replacement of RGB images for classifier training e.g., CNN architectures, which in turn can lead to more accurate classification. Code  will be released for public use.
\end{abstract}

\section*{KEYWORDS}
Image Enhancement, Weighted Least Squares Framework~(WLS), Smoke/Non-Smoke Image Classification.

\section{INTRODUCTION} \label{sec:intro}
\blfootnote{$^{\star}$ Congcong Wang  and  Vivek  Sharma  contributed  equally  to  this  work  and
listed in alphabetical order.}
Over the last decade, we have seen an increase in the number of laparoscopic surgeries~\cite{tsui2013minimally}. During the surgery, such as in cavitary treatment, laser ablation causes smoke~\cite{lawrentschuk2010laparoscopic} which significantly degrades the perceptual quality of the images which inevitably influences the surgeon's visibility, further it also influences the performance of computer vision based navigation systems~\cite{maier2014comparative}. Moreover, surgical smoke is composed of chemical, physical or biological particles, which may be harmful for surgeons and patients~\cite{al2007chemical,choi2014surgical,dobrogowski2014chemical}. Therefore, it is of vital importance to remove the smoke by computer vision algorithms~\cite{wang2018} and by smoke evacuation techniques~\cite{takahashi2013automatic,leibetseder2017image}.  In order to employ a desmoking technique, as a prior knowledge it is essential to know if the image contains smoke or not. In this work, we propose a method to enhance the images for better classification of smoke and non-smoke images.  Our goal is to enhance the images, such that the extracted features from the enhanced images are informative for discrimination that can lead to improved smoke/non-smoke image classification. Note that our goal is not to enhance the images for visual pleasantness of observers' perception, but rather enhance the images features for improved classification.

Our work is inspired from~\cite{sharma_cic,sharma_cvpr}. Sharma~\textit{et al.} in~\cite{sharma_cic} enhance the visible~(RGB) images using near-infrared~(NIR) counterparts and show improvement in the image feature quality for biometric verification tasks, further in~\cite{sharma_cvpr}, Sharma~\textit{et al.} emulates several image enhancement methods in convolutional neural networks for an accurate image classification. We have a similar goal, though our work differs substantially in technical approach and the application scope. Specifically, we utilize weighted least squares optimization framework (WLS)~\cite{ep} to decompose an image to fine and coarse enhanced images, and then combine them in a more meaningful way such that the combined image have better image features for our classification task.

Our proposed approach is evaluated on Cholec80 dataset for smoke/non-smoke image classification. We experimentally show that our proposed method consistently improve the classification performance over the baseline RGB images, popular state-of-the-art enhancement methods, and the saturation histogram based classification methodologies Saturation Analysis (SAN) and Saturation Peak Analysis (SPA).

The remainder of this paper is structured as follows. First, we review the related work on image enhancement and smoke detection methods. Next, we describe our proposed method, and discuss the experimental results. Finally, the conclusions are drawn.

\section{RELATED WORK} \label{sec:related work}

\noindent
\paragraph{\textbf{Image Enhancement.}}
Image enhancement or  filtering techniques enhance the contrast, boost the image details, and produce more vivid colors, and at the same time removes the effects of blur, noise, and compression artifacts. Examples of such filtering methods include weighted least squares (WLS)~\cite{ep}, bilateral filtering~\cite{tomasi}, image sharpening, guided filtering~\cite{ep6}, BFWLS\_AVG~\cite{sharma_cic} and more.   Filtering using (1) RGB and (2) RGB-NIR  images are used for several applications  in computer vision and computational photography applications, such as to improve the contrast of the haze-degraded color images~\cite{dehazing}; tone mapping and detail enhancement~\cite{ep}; denoising in RGB videos~\cite{bf5} and images~\cite{bf9}; multi-modal medical image fusion from MRI-CT~\cite{bf6}; illumination transfer from reference to target images~\cite{bf4}; feature matching~\cite{sharma_feat};  object recognition and image classification tasks~\cite{sharma_cvpr}; biometrics verification tasks~\cite{sharma_cic}; and the list goes on. To the best of our knowledge, our work is the first to show that the image enhancement can be beneficial for smoke/non-smoke image classification. We compare against various of these filtering methods discussed above in our experimental section.

\noindent
\paragraph{\textbf{Smoke Detection.}}
Considerable progress has been seen in the development of ``\textit{in the wild}" video smoke detection techniques over the last decade~\cite{luo2017fire}. The traditional detection methods exploit smoke features, which are intensity, color, motion, and texture  attributes. These features are used to train classifiers for smoke region detection~\cite{kolesov2010fire} or frame detection~\cite{chunyu2010video,calderara2011vision}. It has been found that smoke can reduce the sharpness of edges in order to mitigate over it, several descriptors have been utilized for feature extraction, such as wavelet coefficients~\cite{toreyin2005wavelet}, local binary pattern~(LBP)~\cite{tian2011smoke,yuan2011video}, textural features are estimated from the smoke region only~\cite{tian2014smoke}. Further, in~\cite{tian2018detection}, the authors propose to separate smoke and background from a single image by the dual-dictionary approach, and the estimated sparse coefficients are used as features for smoke frames detection. 

In the medical imaging community, researchers have shown that the formation of smoke and the lighting condition for laparoscopic surgery images are very different, which limits the ``\textit{in the wild}" smoke detection methodologies applicability in medical domain~\cite{leibetseder2017image,leibetseder2017real}. Recently, Chou~\textit{et al.}~\cite{chou2017system} explore temporal differences information that is motion blur and block analysis between current and previous frames as features for smoke detection. Loukas~\textit{et al.}~\cite{loukas2015smoke},  propose a method to detect electrocautery smoke for surgical events retrieval. They extract different features from optical flow estimated by Kanade-Lucas-Tomasi (KLT) algorithm~\cite{shi1994good}, and then a support vector machine (SVM) is trained to classify each shot. Leibetseder~\textit{et al.} in~\cite{leibetseder2017image,leibetseder2017real} propose saturation histogram based classification methodologies Saturation Peak Analysis (SPA) and Saturation Analysis (SAN) and deep learning~(DL) techniques.  In~\cite{leibetseder2017real}, Leibetseder~\textit{et al.} train  three variants of convolutional neural network~(CNN) models: GoogleLeNet~\cite{szegedy2015going} 
trained with RGB images~(GLN RGB) and saturation channels only (GLN SAT), and a modified AlexNet~\cite{krizhevsky2012imagenet} trained on RGB images (ALEX RGB). The performance of their CNN models on Cholec80~\cite{twinanda2017endonet} dataset are very similar to that of SAN or SPA. While DL shows its promising performance on smoke classification~\cite{leibetseder2017image,leibetseder2017real}, to the best of our knowledge there is no work that does image enhancement for smoke classification. Therefore, we limit to comparisons with the non deep learning based methods only in our experimental section. Moreover, one can consistently improve the classification performance by employing our enhancement method and then train the CNN architectures, following the ideas proposed in Sharma~\textit{et al.}~\cite{sharma_cvpr}.

\section{PROPOSED METHOD} \label{sec:proposed method}

\begin{figure}[t]
\centering
\includegraphics[width=0.99\linewidth]{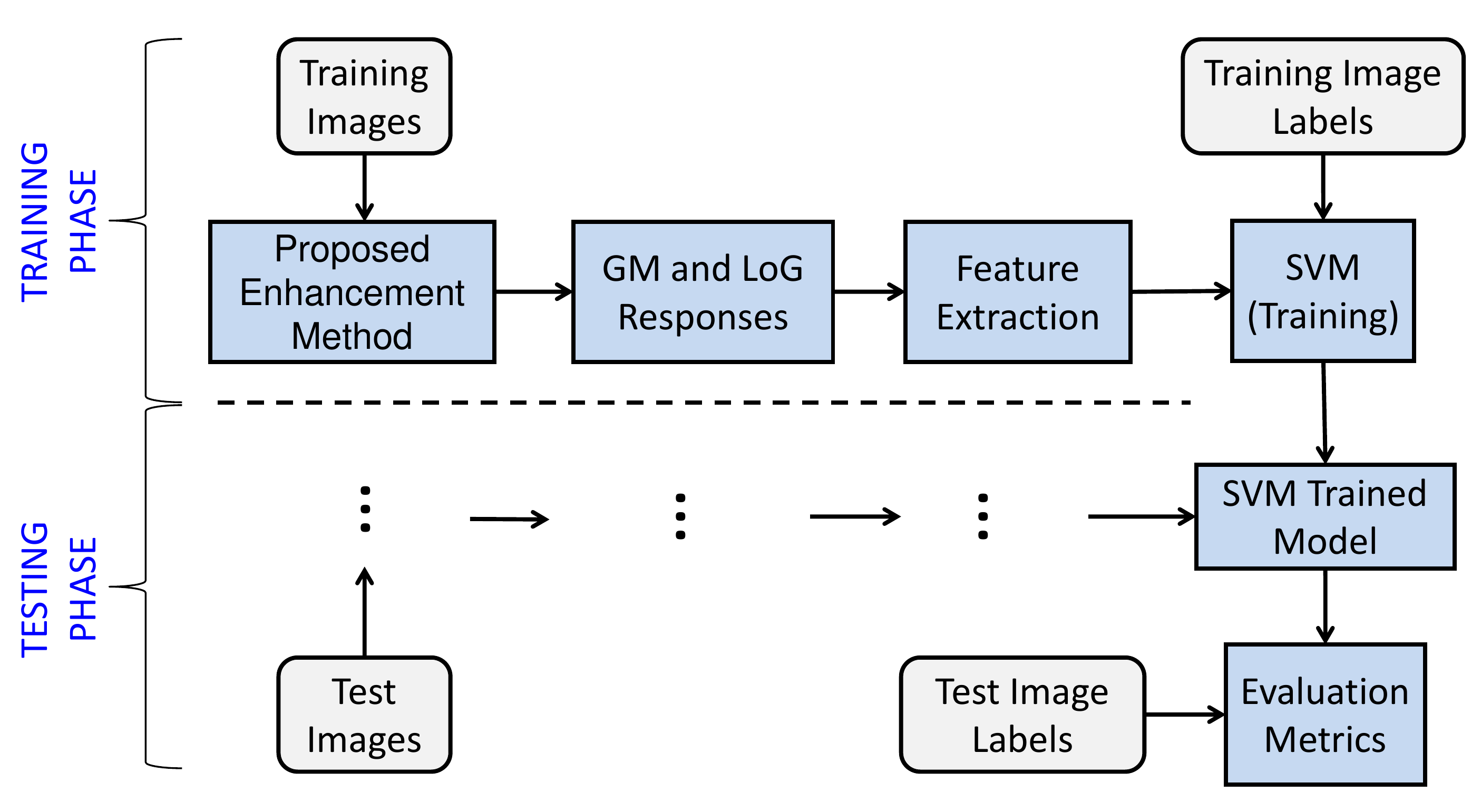}
\vspace{0.25cm}
\caption{Pipeline of the proposed method.}
\label{fig:framework}
\end{figure}

In this section, we illustrate our proposed approach, starting with the proposed enhancement method, then the feature extraction method, and finally classifier training. Figure~\ref{fig:framework} shows the schematic layout of our framework.

\noindent
\paragraph{\textbf{Proposed Enhancement Method.}} We denote our WLS-based filtering method as FC. To  enhance the visible RGB image, we first transform color space from RGB to YCbCr space that is a luminance-chrominance color space. Our method operates on the luminance component~\cite{coloring} such that to modify the overall contrast and sharpness of the image without sensibly affecting the color. Where the chrominance is simply re-combined in the final enhanced image.

In our work, we employ an edge-preserving filter, weighted least squares optimization framework (WLS)~\cite{ep}. The WLS is a non-linear method that captures image details at a variety of scales via multi-scale decompositions. The WLS helps to find an approximate enhanced image $g^{Filtered}$ that is close to the input image $g$, and also at the same time, is smooth along significant gradients, thus resulting to sharper preserved edges. Formally, it is defined as:
\begin{equation}
\hspace{1.5cm}g^{Filtered} = F_{\lambda}(g) = (I +  \lambda L_{g})^{-1}g
\end{equation}
where $L_{g} = D_{x}^{T} A_{x}D_{x} + D_{y}^{T} A_{y}D_{y}$ with $D_{x}$ and $D_{y}$ are discrete differentiation operators.  $A_{x}$ and $A_{y}$ contain the smoothness weights, the smoothness requirement is enforced in a spatially varying manner which depend on $g$. $\lambda$ is the balance factor that maintains a balance between the data term and the smoothness term. Increasing $\lambda$ value  produces progressively smoother images. 

Given an input image, the WLS filter decomposes an image into base and detail layers. The detail layer is simply obtained by subtracting the base layer from the the original image. The base layer comprises of low frequency contents with general appearance of the image over smooth areas, while the detail layer comprises of high frequency contents with sharp edges. 

\begin{figure}[t]
\centering
\includegraphics[width=0.99\linewidth]{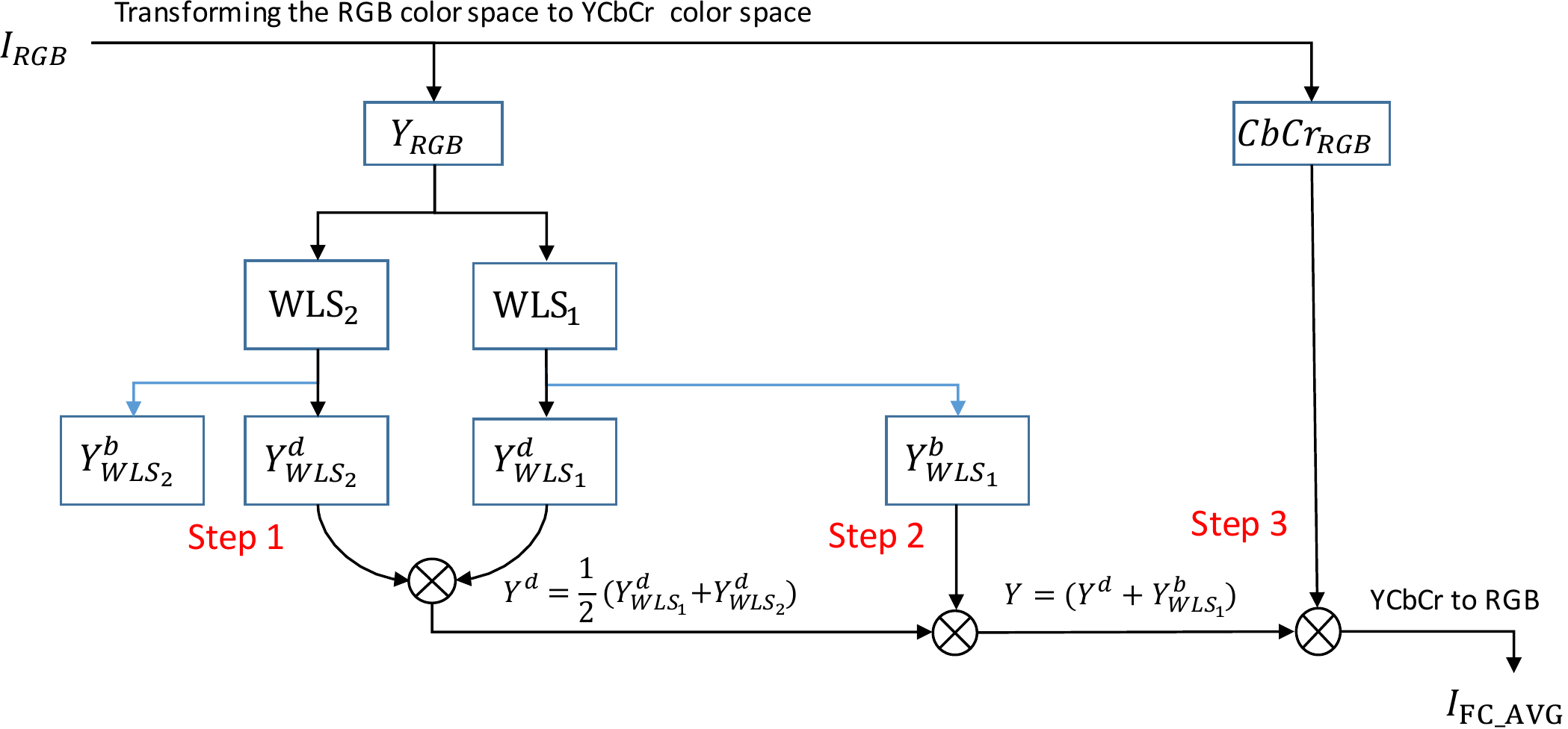}
\vspace{0.25cm}
\caption{Proposed enhancement method.}
\label{fig:enhancement_method}
\end{figure}

In our method, we apply WLS-based two-level decomposition of the luminance component of an RGB image for extraction of fine and coarse enhanced sharp images.  We retain, for each pixel an average value between the fine and coarse detail layers~(Step 1). We chose $\lambda_{1}=0.125$ (WLS$_{1}$), $\lambda_{2}=0.5$ (WLS$_{2}$) for WLS in our experiments. 
The fusion criteria is based on the following observations: the WLS filter is very good at preserving fine and coarse details at arbitrary scales. Taking an average of two, allows to retain the informative content from both, thus allowing to preserve the image structures and also moderately boost the image details. This fusion criteria is denoted as FC\_AVG. In addition, we also tried to retain the maximum values between the two as a fusion criterion. We denote this fusion criteria as FC\_MAX. Finally, we combine the new fused detail layer with the base layer of RGB image obtained using WLS$_{1}$~(Step 2), we consider WLS$_{1}$, as smaller $\lambda$ values are better for the base layer because of less smoothing. And which is then re-combined with the chrominance of RGB image and to reconstruct the final enhanced image~(Step 3). Figure~\ref{fig:enhancement_method} illustrates our proposed method.

\noindent
\paragraph{\textbf{Feature Extraction.}} In our work, we exploit Xue~\textit{et al.}'s method~\cite{xue2014blind} for feature extraction. Xue~\textit{et al.} utilize gradient magnitude (GM) and Laplacian of Gaussian (LoG) maps to describe the structural information for image perceptual quality assessment. Their method is driven by the image statistics, and exploits the histogram information, which is perfect for low-level vision tasks, such as ours. Motivated by this observation, we employ GM and LoG features to represent the local spatial contrast information in images. 

Formally, gradient magnitude~(GM) is defined as:
\begin{equation}
\hspace{1cm}GM=\sqrt{(I\otimes \frac{\partial G}{\partial x})^2+(I\otimes \frac{\partial G}{\partial y})^2},
\end{equation}
where $I$ is the gray scale image, $\otimes$ denotes the convolution operation. $\frac{\partial G}{\partial x}$ and $\frac{\partial G}{\partial y}$ denote the Gaussian partial derivative along $x$ (horizontal) and $y$ (vertical) directions, respectively. They are computed as:

\begin{equation}
\hspace{1.5cm}\frac{\partial G(x,y,\sigma )}{\partial d}=-\frac{1}{2\pi \sigma^{2}}\frac{d}{\sigma^{2}}e^{-\frac{x^2+y^2}{2\sigma^2}},
\end{equation}
where $d \in \{x,y\}$, and $\sigma$ is a scalar parameter. And the Laplacian of Gaussian~(LoG) is defined  as:
\begin{equation}
\centering
\hspace{2cm}LoG=I\otimes h_{LoG}
\end{equation}
\begin{equation}
\begin{split}
h_{LoG}(x,y,\sigma)&=\frac{\partial ^{2} G(x,y,\sigma )}{\partial ^{2} x}+\frac{\partial ^{2} G(x,y,\sigma )}{\partial ^{2} y} \\
&=-\frac{1}{\pi \sigma^{4}}(1-\frac{x^{2}+y^{2}}{2\sigma^{2}})e^{-\frac{x^2+y^2}{2\sigma^2}}.
\end{split}
\end{equation}
Subsequently, a joint adaptive normalization step is applied to adjust the image statistics of the GM and LoG maps, in order to reduce its dependency on local image content. Finally, a bivariate histogram based feature vector is computed from the marginal probability functions and independency distributions of the normalized GM and LoG maps, resulting to a fixed-size feature representation vector of 40 dimensions for an image. We used the default parameters for feature extraction, more details can be found in~\cite{xue2014blind}.

\paragraph{\textbf{Classifier.}} In our work, we use linear SVM~\cite{chang2011libsvm}~\footnote{\texttt{https://www.csie.ntu.edu.tw/$\sim$cjlin/libsvm/}} for a binary classification task in which the objective is to predict the class $y \in \{0, 1\}$  that is  smoke/non-smoke image classification.  We use linear kernel function with C=$10k$ to train/test SVM classifier with the extracted features.

\section{EXPERIMENTS} \label{sec:experiment}

In this section, we demonstrate the use of our proposed enhancement method on a challenging smoke/non-smoke classification dataset~\cite{twinanda2017endonet,leibetseder2017real}. We first introduce the dataset, followed by a thorough analysis of the proposed method. The analysis includes, comparison of our proposed method with baseline RGB images, other popular enhancement methods, and ending with a comparison to saturation histogram based classification methodologies. 

\begin{figure}[ht]
\centering
	\begin{minipage}[b]{0.49\linewidth}
		\centering
		\centerline{\includegraphics[width=1\linewidth]{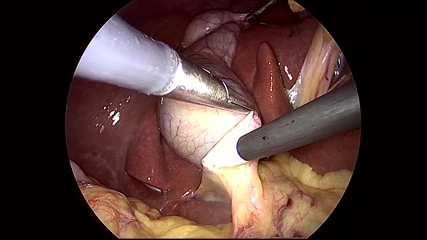}}
		\centerline{ (a) }
	\end{minipage}
	\begin{minipage}[b]{.49\linewidth}
		\centering
		\centerline{\includegraphics[width=1\linewidth]{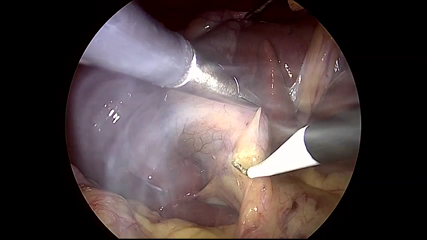}}
		\centerline{ (b) }
	\end{minipage}
    \vspace{0.15cm}
	\caption{Example images from Cholec80 dataset: (a) a smoke free image, (b) image with smoke, covering the surgeon's field of view for visualisation}
	\label{fig:videos}
\end{figure}

\begin{figure*}[ht]

	\begin{minipage}[b]{0.24\linewidth}
		\centering
		\centerline{\includegraphics[width=1\linewidth]{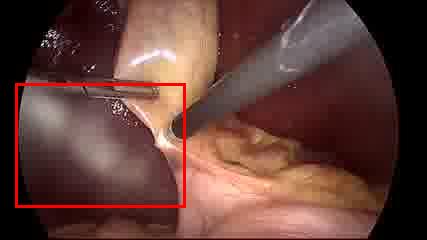}}
		\centerline{ (a) RGB }
	\end{minipage}
	\begin{minipage}[b]{.24\linewidth}
		\centering
		\centerline{\includegraphics[width=1\linewidth]{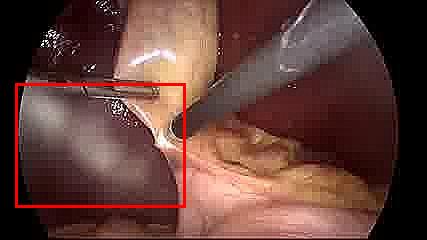}}
		\centerline{ (b) IMSHARP}
	\end{minipage}
	\begin{minipage}[b]{0.24\linewidth}
		\centering
		\centerline{\includegraphics[width=1\linewidth]{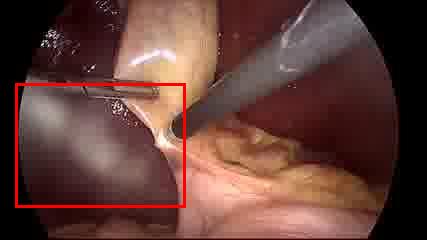}}
		\centerline{ (c) BF}
	\end{minipage}
	\begin{minipage}[b]{0.24\linewidth}
		\centering
		\centerline{\includegraphics[width=1\linewidth]{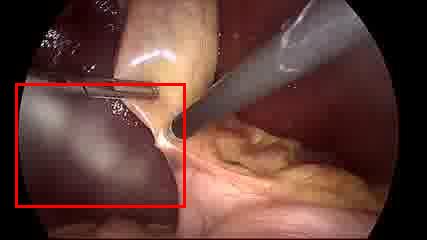}}
		\centerline{ (d) GF}
	\end{minipage}
    
	\begin{minipage}[b]{.24\linewidth}
		\centering
		\centerline{\includegraphics[width=1\linewidth]{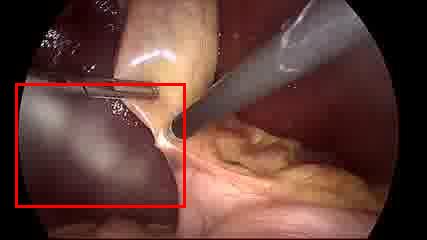}}
		\centerline{ (e) WLS}
	\end{minipage}
	\begin{minipage}[b]{0.24\linewidth}
		\centering
		\centerline{\includegraphics[width=1\linewidth]{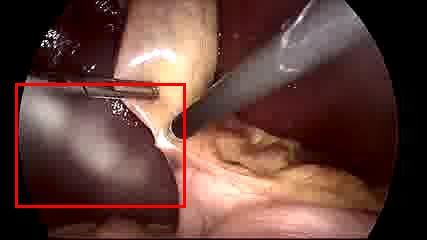}}
        \centerline{ (f) BFWLS\_AVG}
        \end{minipage}
        	\begin{minipage}[b]{0.24\linewidth}
		\centering
		\centerline{\includegraphics[width=1\linewidth]{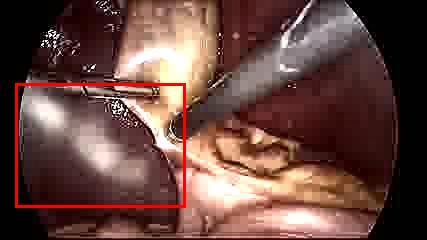}}
        \centerline{ (g) FC\_MAX (\textbf{ours})}
        \end{minipage}
        \begin{minipage}[b]{0.24\linewidth}
		\centering
		\centerline{\includegraphics[width=1\linewidth]{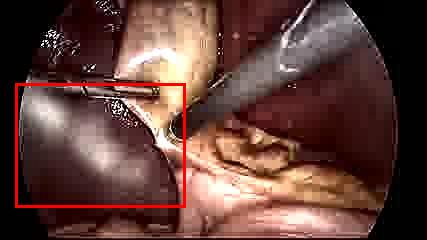}}
		\centerline{ (h) FC\_AVG (\textbf{ours})}
	\end{minipage}    
    \vspace{0.15cm}
    
	\caption{Visual comparison of FC\_AVG and FC\_MAX against other enhancement methods. We can clearly see that the smoke part becomes more perceptually visible after image enhancement. Best viewed in color.}
	\label{fig:enhaced_result}

\end{figure*}

\noindent
\paragraph{\textbf{Dataset.}} We conduct experiments on Cholec80 dataset~\cite{twinanda2017endonet} which contains 80 videos of cholecystectomy surgeries manually labeled with smoke/non-smoke image sequence by~\cite{leibetseder2017real}~\footnote{\texttt{http://www.itec.aau.at/ftp/datasets/Smoke\_cholec80}}. The dataset in overall contains approximately 100K annotated images, in particular between 200-1300 images of smoke/non-smoke in each video. In our experiments, we use a subset of the dataset with three videos for training and nine videos for testing. The videos are randomly chosen. We extract JPEG images, and resize them to a resolution of $427\times240$px from the original resolution of $854\times480$px, to fasten the image enhancement. In particular, we use 4,381 images obtained from video\{1, 31, 34\} for training, and 10,653 images obtained from video\{2, 22, 10, 40, 59, 64, 65, 71, 80\} for testing. In Figure~\ref{fig:videos}, we show some examples of smoke/non-smoke images obtained from Cholec80 dataset. Also, note that each video represents a unique person in this dataset.

\noindent
\paragraph{\textbf{Evaluation Metrics.}} We use accuracy and F1-Score measures computed from a confusion matrix between the predicted labels and the actual ground-truth labels, as the metrics to evaluate the quality of classification.

\noindent
\paragraph{\textbf{Comparison with other enhancement methods.}}
We compare the proposed enhancement method with the baseline RGB images, and popular state-of-the-art enhancement approaches: (1) guided filtering (GF)~\cite{ep6}, (2) bilateral filtering (BF)~\cite{tomasi,fbf}, (3) image sharpening filter (IMSHARP), (4) WLS~\cite{ep}, (5) fused BF and WLS filter~(BFWLS\_AVG)~\cite{sharma_cic}. In BFWLS\_AVG, Sharma~\textit{et al.}~\cite{sharma_cic} apply BF and WLS filters on the NIR channel, and retain an average of two for each pixel, similar to~\cite{sharma_cic}, instead we employ their technique in the RGB image. For a fair comparison, we compare all the methods under the same evaluation protocol discussed above. For the evaluation, we use the same parameters for feature extraction and classifier training for all methods. We did not optimize the parameters for the BF, GF, WLS, IMSHARP, and BFWLS\_AVG and used the default parameters for each method. The source code for fast BF~\footnote{\texttt{http://people.csail.mit.edu/sparis/bf/}}, WLS~\footnote{\texttt{http://www.cs.huji.ac.il/$\sim$danix/epd/}} and BFWLS\_AVG~\footnote{\texttt{https://vivoutlaw.github.io/CIC\_RGB\_NIR\_Codes.zip}} are publicly available, and others are available in the Matlab framework. For comprehensive discussion of different methods, we refer the reader to~\cite{fbf,ep,ep6,sharma_cic}.

\begin{figure*}[htb]
\centering
	\begin{minipage}[b]{0.3\linewidth}
		\centering
		\centerline{\includegraphics[width=1\linewidth]{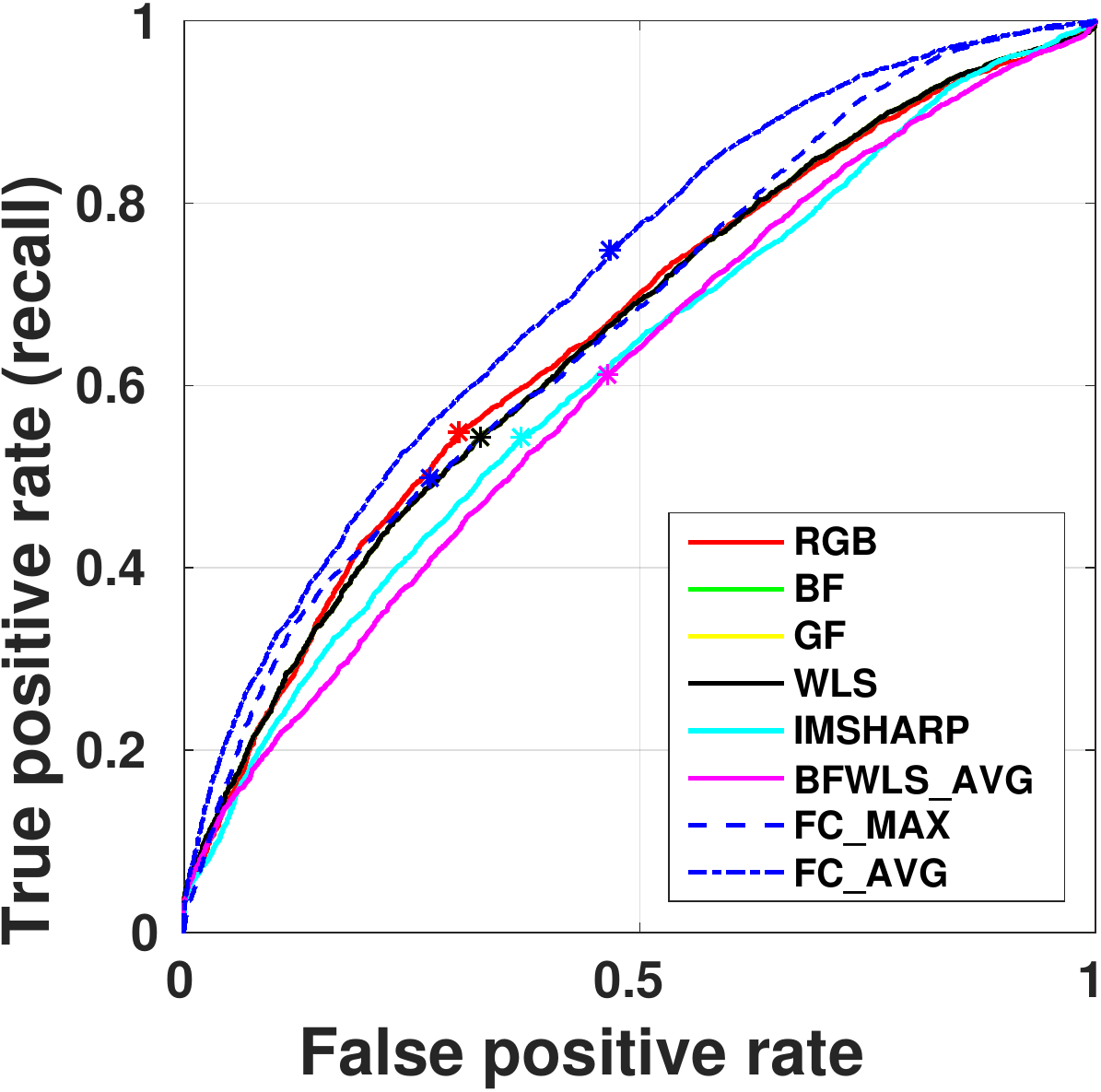}}
		\centerline{ (a) Comparison with other enhancement methods.}
	\end{minipage}
    \qquad \qquad
	\begin{minipage}[b]{.3\linewidth}
		\centering
		\centerline{\includegraphics[width=1\linewidth]{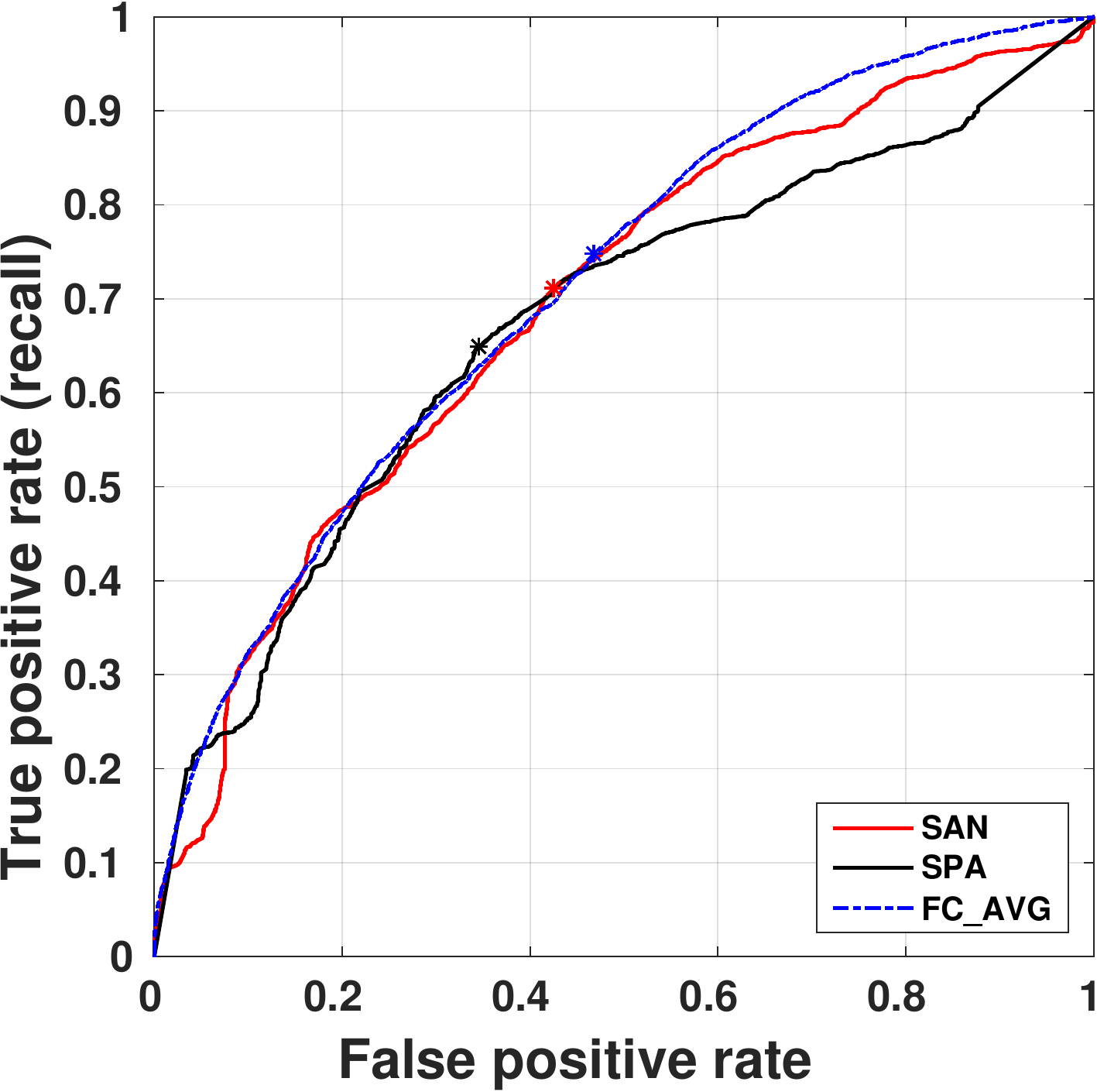}}
		\centerline{ (b) Comparison with SAN and SPA.}
	\end{minipage}
    \vspace{0.25cm}
	\caption{The ROC curves for the Cholec80 dataset using GM-LoG features~\cite{xue2014blind} for smoke/non-smoke classification task. \textbf{*} denotes the EER when the false accept rate is equal to the false reject rate. Best viewed in color.}
	\label{fig:roc}
\end{figure*}

In Table~\ref{tab:result1}, we quantitatively evaluate the accuracy and F1-Score of our proposed method and other methods. We can clearly observe that  FC\_AVG performs the best among all methods. Enhancing an image using FC\_AVG takes approximately 0.32 seconds. Note that, FC\_AVG improves over the baseline performance of RGB images by  4\% in accuracy and 4\% in F1-Score, giving higher-quality features to learn from, which in turn lead to more accurate classification. We believe our work opens many possibilities for further exploration for its usage in other tasks too. Further the performance gap of FC\_AVG is 4/5\% better than FC\_MAX in accuracy/F1-Score measures. Our method FC\_AVG consistently performs better than all other methods: IMSHARP, BF, WLS, and GF enhancement methods. In addition, our methods also performs better than BFWLS\_AVG, although we agree that BFWLS\_AVG is meant for RGB-NIR fusion and not RGB image enhancement, that may be the reason why it underperforms to our method. Figure~\ref{fig:roc}~(a) shows the ROC curve for all the methods. 

In Figure~\ref{fig:enhaced_result}, we compare enhancement methods for an example image obtained from video1. We can observe that FC\_AVG has improved features for the smoke-part (highlighted by red rectangles), in addition to reduced specular intensity when compared to FC\_MAX. We believe this plays an important role in discrimination.

\begin{table}[htb]
\tabcolsep=0.3cm
\begin{center}
\begin{tabular}{lcc}
\toprule
Method & Accuracy &F1-Score \\ 
\midrule
\midrule
RGB 		& 0.60& 0.60\\
\midrule
IMSHARP		& 0.58& 0.58\\
BF~\cite{fbf}			& 0.60& 0.59\\
GF~\cite{ep6}			& 0.60& 0.59\\
WLS~\cite{ep}			& 0.60& 0.59\\
\midrule
BFWLS\_AVG~\cite{sharma_cic} & 0.57& 0.56\\
\midrule
FC\_MAX~(\textbf{Ours})  	& 0.60& 0.59\\
FC\_AVG~(\textbf{Ours}) 	& \textbf{0.64}& \textbf{0.64}\\
\bottomrule
\end{tabular}
\vspace{0.25cm}
\caption{Comparison with the baseline RGB images and other enhancement methods.}
\label{tab:result1}
\end{center}
\end{table}

\noindent
\paragraph{\textbf{Comparison with the saturation histogram based classification methodologies.}}
We compare FC\_AVG with the saturation histogram based classification methodologies Saturation Analysis (SAN) and Saturation Peak Analysis (SPA)~\cite{leibetseder2017image,leibetseder2017real}. SAN and SPA codes are provided by~\cite{leibetseder2017image,leibetseder2017real}~\footnote{\texttt{https://github.com/amplejoe/SaturationPeakAnalysis}}. SAN and SPA take advantage of the saturation channel of the HSV color space. In~\cite{leibetseder2017image,leibetseder2017real}, the authors show that the histogram bin curve of saturation channel is strongly correlated with the presence of smoke, as smoke images contain more low-saturation pixels and that is helpful for classifying smoke/non-smoke images. In their method, a threshold $t_{c}$ is set empirically. For SAN method, smoke image is recognized if majority of the bin values are below $t_{c}$. For SPA method, smoke image is classified by the number of peaks found below and above $t_{c}$. We use the default parameters suggested by Leibetseder~\textit{et al.}~\cite{leibetseder2017image,leibetseder2017real} for both SAN and SPA where $t_c$ is set to 0.35.

It is evident from Table~\ref{tab:result2} that FC\_AVG shows improved performance over the saturation histogram based classification methodologies: SAN and SPA. Precisely, FC\_AVG is 1/5\% and 1/6\% better than the SAN and SPA methods in accuracy/F1-Score metrics. Further in Figure~\ref{fig:roc}~(b), we show the ROC curve for FC\_AVG, SAN and SPA classification methodologies.

\begin{table}[htb]
\tabcolsep=0.3cm
\begin{center}
\begin{tabular}{lcc}
\toprule
Method & Accuracy &F1-Score \\ 
\midrule
\midrule
SPA~\cite{leibetseder2017image} &0.63 &0.58 \\
SAN~\cite{leibetseder2017real} &0.63 & 0.59\\
\midrule
FC\_AVG~(\textbf{Ours}) 	& \textbf{0.64}& \textbf{0.64}\\
\bottomrule
\end{tabular}
\vspace{0.25cm}
\caption{Comparison with the saturation histogram based classification methodologies Saturation Analysis~(SAN) and Saturation Peak Analysis~(SPA)}
\label{tab:result2}
\end{center}
\end{table}

\section{CONCLUSION} \label{sec:conclusion}
We present a method to enhance RGB images using weighted least squares filter. Our method successfully enhances the informative features in the images for discrimination of smoke/non-smoke images. We demonstrate our proposed method on Cholec80 dataset. Our proposed method obtains significant gains over the baseline RGB images and other enhancement methods. Our method also outperforms the saturation histogram based classification methodologies. Our enhancement method can be used to perform explicit enhancement of image texture and structure features for other tasks too which in turn can lead to more improved performance.

In future work, we plan to further investigate our method on the complete Cholec80 dataset, and also train a CNN architecture with our enhanced RGB images.




\bibliographystyle{IEEEtran}
\bibliography{main}

\end{document}